\documentclass[10pt,twocolumn,letterpaper]{article}

 \usepackage[pagenumbers]{cvpr} 
\usepackage{algorithm}
\usepackage{algpseudocode}
\usepackage{xcolor}
\usepackage{listings}
\usepackage{enumitem}

\algrenewcommand\algorithmicrequire{\textbf{Input:}}
\algrenewcommand\algorithmicensure{\textbf{Output:}}
\floatname{algorithm}{Masking Strategy}

\newlist{insightlist}{enumerate}{1}
\setlist[insightlist]{
    label=\textbf{Insight \arabic*.},
    leftmargin=*,
    itemsep=0.5em
}

\definecolor{cvprblue}{rgb}{0.21,0.49,0.74}
\usepackage[pagebackref,breaklinks,colorlinks,allcolors=cvprblue]{hyperref}
\usepackage{svg}
\def\paperID{**} 
\def\confName{}
\def\confYear{2026}

\title{Hide and Seek: Investigating Redundancy in Earth Observation Imagery}
\author{Tasos Papazafeiropoulos$^{1\ast}$ \and Nikolaos Ioannis Bountos$^{2\ast\S}$ \and Nikolas Papadopoulos$^{1}$ \and Ioannis Papoutsis$^{1}$ \\
$^1$ Orion Lab\\
  National Observatory of Athens \& National Technical University of Athens\\
$^2$ Independent Researcher
}

\begin{document}
\maketitle 
\begin{abstract}
The growing availability of Earth Observation (EO) data and recent advances in Computer Vision have driven rapid progress in machine learning for EO, producing domain-specific models at ever-increasing scales. Yet this progress risks overlooking fundamental properties of EO data that distinguish it from other domains. We argue that EO data exhibit a multidimensional redundancy (spectral, temporal, spatial, and semantic) which has a more pronounced impact on the domain and its applications than what current literature reflects. To validate this hypothesis, we conduct a systematic domain-specific investigation examining the existence, consistency, and practical implications of this phenomenon across key dimensions of EO variability. Our findings confirm that redundancy in EO data is both substantial and pervasive: exploiting it yields comparable performance ($\approx98.5\%$ of baseline) at a fraction of the computational cost ($\approx4\times$ fewer GFLOPs), at both training and inference. Crucially, these gains are consistent across tasks, geospatial locations, sensors, ground sampling distances, and architectural designs; suggesting that multi-faceted redundancy is a structural property of EO data rather than an artifact of specific experimental choices. These results lay the groundwork for more efficient, scalable, and accessible large-scale EO models.
\end{abstract}
\renewcommand{\thefootnote}{}
\footnotetext{$^\ast$Equal contribution.}
\footnotetext{$^\S$Work done while at the Orion Lab.}
\section{Introduction}
\label{sec:intro}

\begin{figure}[t]
  \centering
   \includegraphics[width=\linewidth]{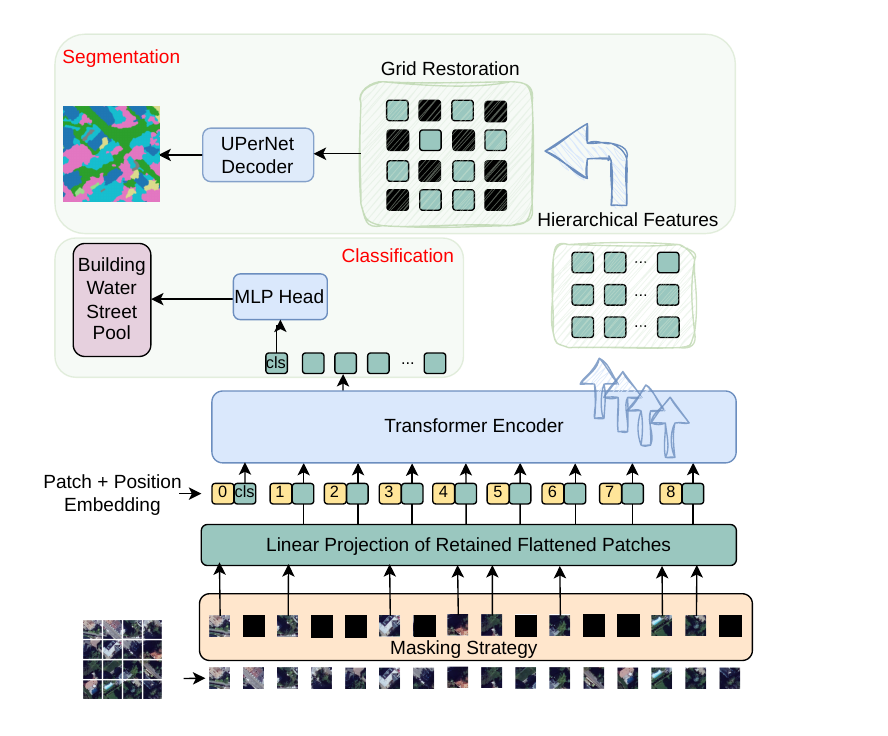}

   \caption{Overview of RViT and RViT-UperNet for classification and semantic segmentation.}
   \label{fig:architecture}
\end{figure}


The growing availability of Earth Observation (EO) data has enabled the community to exploit advances in Deep Learning for critical EO 
applications such as land cover classification \cite{reben}, tree species identification\cite{kattenborn2019uav}, and marine debris detection\cite{kikaki2022marida}. The global coverage of satellite missions and the broad applicability of EO data have driven the development of large-scale EO foundation models, capable of operating across diverse sensors, tasks, locations, and environmental conditions~\cite{bountos2025fomo,wang2025towards,tseng2023lightweight,
fuller2023croma,cong2022satmae}. The prevailing approach for constructing such models is Masked Image Modelling (MIM), primarily building on 
Masked Autoencoders (MAE) ~\cite{he2022masked}, where large portions of the image are masked and reconstructed as a self-supervised pretraining objective. 
MAE was designed for natural images where spatial redundancy between neighboring pixels makes masked reconstruction a meaningful yet challenging task. EO imagery, however, is fundamentally different from natural images~\cite{kondylatos2025generalization,
rolf2024position}. It encompasses extreme sensor and spectral diversity, varying environmental conditions and topography, highly diverse object 
scales and granularities, and ground sampling distances (GSDs) spanning from a few centimeters to kilometers. Moreover, unlike natural images, EO imagery does not have a clear notion of background, as most regions of a scene carry semantic information that could potentially be considered a target. These properties, along with the success of MIM-based models in EO, lead us to a hypothesis that redundancy, in all its multifaceted nature (temporal---repetetive patterns through time, spatial---repetitive and continuous regions, spectral---overlapping information across adjacent or nearby wavelength bands, and conceptual/semantic---repetition of semantic concepts in arbitrary locations) has a more pronounced impact on the domain than current literature reflects. 
The above give rise to the following question: \textit{Is redundancy an inherent property of EO imagery across its factors of variability, and if so, how does it manifest and how reliably can we exploit it for efficiency and performance?} 
To investigate this, we build on the Semantic Factors (SFs) of EO variability introduced by \cite{kondylatos2025generalization} and construct a systematic investigation framework that examines redundancy across each of these axes (see \cref{sec:datasets}). We 
introduce redundancy-aware Vision Transformers (ViT) \cite{dosovitskiy2020vit}, \ie, \textit{RViT} and \textit{RViT-UPerNet}, that employ redundancy reduction via task-agnostic image-space patch masking, enabling major computational cost reduction with minimal performance drop and improved sustainability. We examine three complementary redundancy reduction strategies with distinct underlying intuitions, from assumption-free uniform random patch selection, to diversity-based patch selection, and study their behavior across a wide array of classification and segmentation tasks. Redundancy reduction enables models to retain the vast majority of their predictive skill at a fraction of the computational cost: retaining only $25\%$ of patches in MLRSNet (classification) yields $>98.5\%$ of baseline performance at $\approx4\times$ fewer GFLOPs, while on the high-resolution FLAIR (segmentation) dataset we recover $\approx97\%$ of baseline performance at $\approx2\times$ fewer GFLOPs, in both training and inference. 
The current study focuses on spatial redundancy; cross-channel and temporal redundancy remain compelling directions for future work. 
Our study uncovers the following key insights:
\begin{itemize}
    \item The existence of redundancy as a property of EO imagery is substantial and pervasive across all SFs of variability.
    \item The impact of redundancy reduction is robust against the choice of masking strategy.
    \item EO tasks remain solvable with a fraction of available information, even at inference.
    \item Redundancy reduction retains predictive skill with majorly improved efficiency.
\end{itemize}

\section{Related Work}
\label{sec:related_word}


\subsection{Data Pruning}

Data pruning is a well investigated field in Deep Learning that seeks to identify and discard redundant or noisy samples \cite{xia2022moderate,tan2023data}. 
Recent works have adopted this concept for the EO domain. \citet{Kerdreux_2025_CVPR} introduced a method that enhances the diversity of pretraining datasets for self-supervised learning. Similarly, \cite{wei2025rs} introduced RS-Prune, a training free method to select a subset of a dataset for training foundation models. Conceptually, our masking approach is related to \emph{data pruning}. However, it differs fundamentally from conventional instance-level pruning techniques that remove entire samples. 
Our method operates \emph{within} each image, pruning patches rather than instances, which can be directly applied to inference contexts. 
\subsection{Token redundancy, merging and pruning}
Prior works have studied the various forms of redundancy that arise in ViTs \cite{chen2022principle}. One such source  is patch embedding redundancy, \ie redundancy at the token space. To address this, the authors proposed to diversify tokens using regularization terms enforcing within-layer and cross-layer diversity. A related line of work addresses token-level redundancy through pruning and merging based on similarity and importance scores \cite{rao2021dynamicvit,liang2022not,bolya2022token}. Building on this progress, recent works adopted token merging in EO. \citet{Niu2025ATMformerAA} introduced ATMFormer that employs token merging for EO scene classification. \citet{Huang2024HyperspectralIC} employed token fusion to reduce computational costs in hyperspectral image classification. Finaly, \citet{Chen2025SPRINTAM} introduced SPRINT, a pretraining framework utilizing a token reduction strategy, by merging consecutive similar tokens, and removing the tokens with lowest attention scores. These methods demonstrate that redundancy can be exploitable in image data; yet, none of of them examines how redundancy manifests in EO, nor do they investigate its structure across domain-specific variability axes. Our work addresses this gap directly.

\subsection{Masked Image Modelling}
Masking has been extensively utilized for large-scale self-supervised learning in EO foundation models. MAE \cite{he2022masked} introduced an efficient masked reconstruction pretraining paradigm for natural images, which many EO approaches have since adapted. SatMAE \cite{cong2022satmae} was among the first to apply MIM to EO data, while Scale-MAE~\cite{reed2023scale} extended it to account for varying GSDs. FoMo-Net \cite{bountos2025fomo} proposed a flexible MIM pretraining framework supporting the most commonly used EO modalities and spectral bands within a single backbone. Presto~\cite{tseng2023lightweight} introduced a MIM framework for pixel-level time series that is able to process a varying number of spectral bands. CopernicusFM \cite{wang2025towards} further extended MIM to flexibly incorporate spectral, non-spectral, and metadata information including geolocation and acquisition time. While all of the above implicitly exploit spatial redundancy as a pretraining signal, none investigate its impact as an explicit domain property. 


\section{Investigation Framework}

This section outlines our investigation framework, describing
the methodology (\cref{sec:methodology,sec:masking_strategies}), and examined datasets and investigation principles (\cref{sec:datasets}). 


\subsection{RViT: Redundancy-aware Vision Transformer}
\label{sec:methodology}

We examine the existence of redundancy and its implications by utilizing an image-space masking mechanism that selectively retains patches during the forward pass following a predefined masking strategy \textit{MS}. We focus on ViTs for this investigation due to their widespread adoption in large-scale EO foundation models and their sequence nature---enabling full computational savings from patch omission without architectural modifications. Formally, each sample $\mathcal{I} \in \mathbb{R}^{H \times W \times C}$ is partitioned into $N = (H\times W)/P^2$ non-overlapping patches of size $P$. According to a given masking strategy, we select a subset of patch indices $\mathcal{S} \subseteq \{1, \dots, N\}$ of size $N_{r}<N$, which are linearly projected to a dimension $D$ and fed to the Transformer. The remaining $N-N_{r}$ patches are discarded entirely, yielding direct reductions in memory and compute, amplified by the quadratic complexity of self-attention with respect to the sequence length.
For classification the prediction is obtained directly from the classification token via a linear head. For semantic segmentation, we extract hierarchical token representations from four intermediate Transformer blocks $b_\ell$ for $\ell \in \{1\dots4\} $, yielding stage-wise token features:
\[
\mathbf{F}_r^\ell = 
\{\mathbf{f}_1^\ell, \dots, \mathbf{f}_{N_r}^\ell\}
\in \mathbb{R}^{N_r \times D},
\]

where $f^l_i$ denotes the representation of token $i$ obtained from the transformer block $l$. To recover the full spatial structure for dense prediction, the retained tokens are scattered back to their original spatial positions, with masked locations filled with zero-tokens as neutral placeholders. This restores the original grid topology while enabling redundancy-aware selection. The final dense token sequence is then reshaped into spatial feature maps $\mathbf{F}^\ell \in \mathbb{R}^{D \times H_\ell \times W_\ell}$, for $\ell \in \{1\dots4\}$ and fed to an UPerNet decoder \cite{xiao2018unified} for the segmentation prediction. The full pipeline can be seen in \cref{fig:architecture}. We refer to the classification and segmentation variants as \textit{RViT} and \textit{RViT-UPerNet}.


\subsection{Masking strategies}
\label{sec:masking_strategies}
RViT and RViT-UPerNet rely on a patch selection mechanism to omit redundant patches. Since there is no canonical definition for identifying and quantifying redundancy in image regions, we examine three complementary masking strategies to assess how the definition of redundancy reduction alters the effect of the phenomenon. 
We consider: (\textit{MS1}) \textit{uniform random} masking, (\textit{MS2}) \textit{diversity-based} masking, and (\textit{MS3}) \textit{thresholded diversity-based} masking. \textbf{\textit{MS1}} enforces spatially uniform sparsification by randomly masking patches, encouraging the model to operate under evenly distributed information removal and thus exploiting spatial autocorrelation (\cref{alg:ms1}). \textbf{\textit{MS2}} retains a fixed proportion of the most diverse patches, where diversity is quantified via cosine similarity in pixel space, thereby prioritizing patches that contribute maximal variance (\cref{alg:ms2}). \textbf{\textit{MS3}} extends this principle by dynamically determining the retention ratio based on a diversity threshold, allowing the sparsity level to adapt to scene-specific redundancy (\cref{alg:ms3}).  We analyze all strategies across training and inference, on classification and segmentation tasks. Visual comparison between the aforementioned masking strategies is available in Suppl \cref{sec:viz_mask}. 




\begin{algorithm}[t]
\caption{Uniform Random Patch Masking}
\label{alg:ms1}
\begin{algorithmic}[1]
\Require Sample key $\kappa$, 
         retention ratio $r \in (0,1]$,
         total token count $N$, hash function $f$
\Ensure  Binary mask $\mathbf{M} \in \{0,1\}^N$,  $\sum_i M_i = k$

\State $k \leftarrow \lfloor r \cdot N \rfloor$

\State $\text{seed} \leftarrow f(\kappa)$ \Comment{Deterministic per-sample seed}

\State $\mathcal{G} \leftarrow \texttt{manual\_seed}(\text{seed})$

\State $\pi \leftarrow \texttt{randperm}(N,\, \mathcal{G})$\Comment{Seeded random permutation}

\State $\mathcal{I} \leftarrow \pi_{1:k}$\Comment{Select first $k$ indices}

\For{$i = 1, \dots, N$}
    \State $M_i \leftarrow \mathbf{1}[i \in \mathcal{I}]$
\EndFor

\State \Return $\mathbf{M}$
\end{algorithmic}
\end{algorithm}

\begin{algorithm}[t]
\caption{Diversity-Based Patch Masking}
\label{alg:ms2}
\begin{algorithmic}[1]
\Require Similarity matrix $\mathbf{S} \in \mathbb{R}^{N \times N}$, 
         retention ratio $r \in (0,1]$
\Ensure  Binary mask $\mathbf{M} \in \{0,1\}^N$, $\sum_i M_i = k$ 

\State $k \leftarrow \lfloor r \cdot N \rfloor$

\For{$i = 1, \dots, N$}
    \State $\bar{s}_i \leftarrow \frac{1}{N} \sum_{j=1}^{N} S_{i,j}$
\EndFor

\State $\mathcal{I} \leftarrow \bigl\{\arg\min_{i}\, \bar{s}_i\bigr\}$ 
\Comment{Init. with most dissimilar token}

\While{$|\mathcal{I}| < k$}
    \For{$i \notin \mathcal{I}$}
        \State $d_i \leftarrow \max_{j \in \mathcal{I}}\, S_{i,j}$
        \Comment{Max similarity to $\mathcal{I}$}
    \EndFor
    \State $\mathcal{I} \leftarrow \mathcal{I} \cup 
           \bigl\{\arg\min_{i \notin \mathcal{I}}\, d_i\bigr\}$
    \Comment{Add most dissimilar candidate}
\EndWhile
\State $M_i \leftarrow \mathbf{1}[i \in \mathcal{I}], \quad \forall i \in \{1,\dots,N\}$
\Comment{Construct binary mask}
\State \Return $\mathbf{M}$
\end{algorithmic}
\end{algorithm}

\begin{algorithm}[t]
\caption{Thresholded Diversity-Based Patch Masking}
\label{alg:ms3}
\begin{algorithmic}[1]
\Require Similarity matrix $\mathbf{S} \in \mathbb{R}^{N \times N}$,
         similarity threshold $\tau \in (0,1]$,
         batch size $B$, token embeddings $\mathbf{F} \in \mathbb{R}^{B \times N \times D}$
\Ensure  Tokens $\{\mathbf{V}_b\}_{b=1}^B$,
         attention masks $\{\mathbf{A}_b\}_{b=1}^B$

\For{$b = 1, \dots, B$}
    \For{$i = 1, \dots, N$}
        \State $M_i^b \leftarrow \mathbf{1}\bigl[\max_{j \neq i}\, S_{i,j}^b < \tau\bigr]$\Comment{Token $i$ informative if dissimilar to all others}
    \EndFor
    \State $\mathcal{S}_b \leftarrow \{i : M_i^b = 1\}$
    \Comment{Retained set for sample $b$}
    \State $k_b \leftarrow |\mathcal{S}_b|$
\EndFor

\State $\texttt{seq\_len} \leftarrow \min\bigl(\max_b\, k_b,\, N\bigr)$\Comment{Batch seq. length}

\For{$b = 1, \dots, B$}
    \State $\mathbf{V}_b \leftarrow \bigl[\mathbf{f}_i^b : i \in \mathcal{S}_b\bigr]$
    \Comment{Retained tokens}
    \State $\mathbf{V}_b \leftarrow \texttt{pad}\bigl(\mathbf{V}_b,\, \texttt{seq\_len}\bigr)$\Comment{Pad with zero}
    \State $A_b^i \leftarrow \mathbf{1}[i \leq k_b], \quad \forall i \in \{1,\dots,\texttt{seq\_len}\}$ 
\EndFor

\State \Return $\{\mathbf{V}_b\}_{b=1}^B$, $\{\mathbf{A}_b\}_{b=1}^B$
\end{algorithmic}
\end{algorithm}

\subsection{Investigation Principles \& Datasets}
\label{sec:datasets}

\noindent\textbf{Investigation Principles.} 
We build our investigation on \cite{kondylatos2025generalization}, which identified core properties of EO data and defined four \textit{Semantic Factors} (SFs) of variability that capture the universal modality-invariant concepts of
EO imagery: \textbf{SF1: Ground Sampling Distance}---the spatial resolution of each sensor, \textbf{SF2: Domain of Interest}---the geographic and thematic factors, \textbf{SF3: Granularity of the Targets}---the semantic detail across EO tasks, and \textbf{SF4: Spatial arrangement}---the spatial layout of objects in a scene. We adopt this concept and construct an investigation framework studying redundancy across 
these SFs. 
\begin{table*}[t]
\label{tab:datasets}
\centering
\resizebox{\textwidth}{!}{%
\begin{tabular}{l l l l c l c c c c c}
\toprule
\textbf{Dataset} & \textbf{Input Modality} & \textbf{Sensor} & \textbf{DL Task} & \textbf{\# Classes} & \textbf{EO Task} & \textbf{Spatial Res.} & \textbf{Image Size} & \textbf{Coverage} \\
\midrule
BigEarthNet & MS/SAR & S1, S2 & Multi-label classification & 19 & LULC classification & 10m & $120 \times 120$ & Europe \\
BigEarthNet-5 & MS/SAR & S1, S2 & Multi-label classification & 5 & LULC classification & 10m & $120 \times 120$ & Europe \\
MLRSNet & RGB & Multi-sensor & Multi-label classification & 60 & Semantic Scene Understanding & $\approx$10 - 0.1m & $256 \times 256$ & Global \\
Woody & RGB & Aerial & Image Segmentation & 4 & Tree-species detection & 50cm & $224 \times 224$ & Chile \\
Waititu & RGB & Aerial & Image Segmentation & 3 & Invasion tree-species detection & 50cm & $224 \times 224$ & New Zealand \\
Flair & RGB/NIR/DEM & Aerial & Image Segmentation & 19 & LULC semantic segmentation & 20cm & $512 \times 512$ & France \\
MARIDA & MS & S2 & Image Segmentation & 12 & Marine Debris Detection & 10m & $224 \times 224$ & Global \\
\bottomrule

\end{tabular}%
}
\caption{Taxonomy of datasets examined in this study. S1/S2 stand for Sentinel-1/2, MS for Multispectral, SAR for Synthetic Aperture Radar) and NIR for Near Infrared. DEM stands for Digital Elevation Model and LULC for Land Use Land Cover. }
\label{tab:datasets}
\end{table*}

\noindent\textbf{Datasets.} Taking the above into consideration we collect a diverse set of datasets (see \cref{tab:datasets}) capturing wide properties of EO data. 
\textit{BigEarthNetV2} \cite{reben} is a multi-label classification dataset revolving around Land Use and Land Cover classification, comprising Synthetic Aperture Radar and Multispectral data from the Sentinel-1 and Sentinel-2 missions over Europe at 10m per pixel. We will refer to this dataset as BigEarthNet. \textit{BigEarthNet-5} is a variant of BigEarthNet introduced in \cite{kondylatos2025generalization} by merging the initial 19 classes into 5 coarser object categories directly affecting target granularity. \textit{MLRSNet} \cite{mlrsnet} comprises data from varying sensors and GSDs (10m to 10cm), with global spatial coverage, addressing the problem of scene understanding. \textit{Woody}~\cite{kattenborn2019uav} and \textit{Waititu}~\cite{kattenborn2020convolutional} are high-resolution UAV datasets, offering data at 50cm per pixel, addressing the tree species segmentation problem in Chile and New Zealand, respectively. \textit{Flair}~\cite{garioud2023flair,ign2022flair1} is a high-resolution aerial dataset at a 20cm per pixel spatial resolution for LULC semantic segmentation across France. Finally, \textit{MARIDA}~\cite{kikaki2022marida} is a global dataset for marine debris detection in coastal and riverine environments, offering multispectral data from the Sentinel-2 mission at a 10m per pixel spatial resolution.
\section{Experiments \& Discussion}
We base our experiments on three ViT variants: ViT-Small, ViT-Base, and ViT-Large and initialize them with weights pretrained on ImageNet \cite{deng2009imagenet}. We use all available channels for each dataset unless stated otherwise. For each patch, all channels are aggregated in one token, leaving cross-channel redundancy as future work. Notably, MS3 operates on similarity thresholds resulting in dynamic patch sequences, and is thus not directly comparable to retention ratios presented for MS1 and MS2. Suppl \ref{sec:suppl_correspondence} presents correspondence of similarity threshold to retention ratios per dataset.
\subsection{Characterizing redundancy in EO imagery}
\label{sec:existence}
\begin{figure*}
    \centering
    \includegraphics[width=0.95\linewidth]{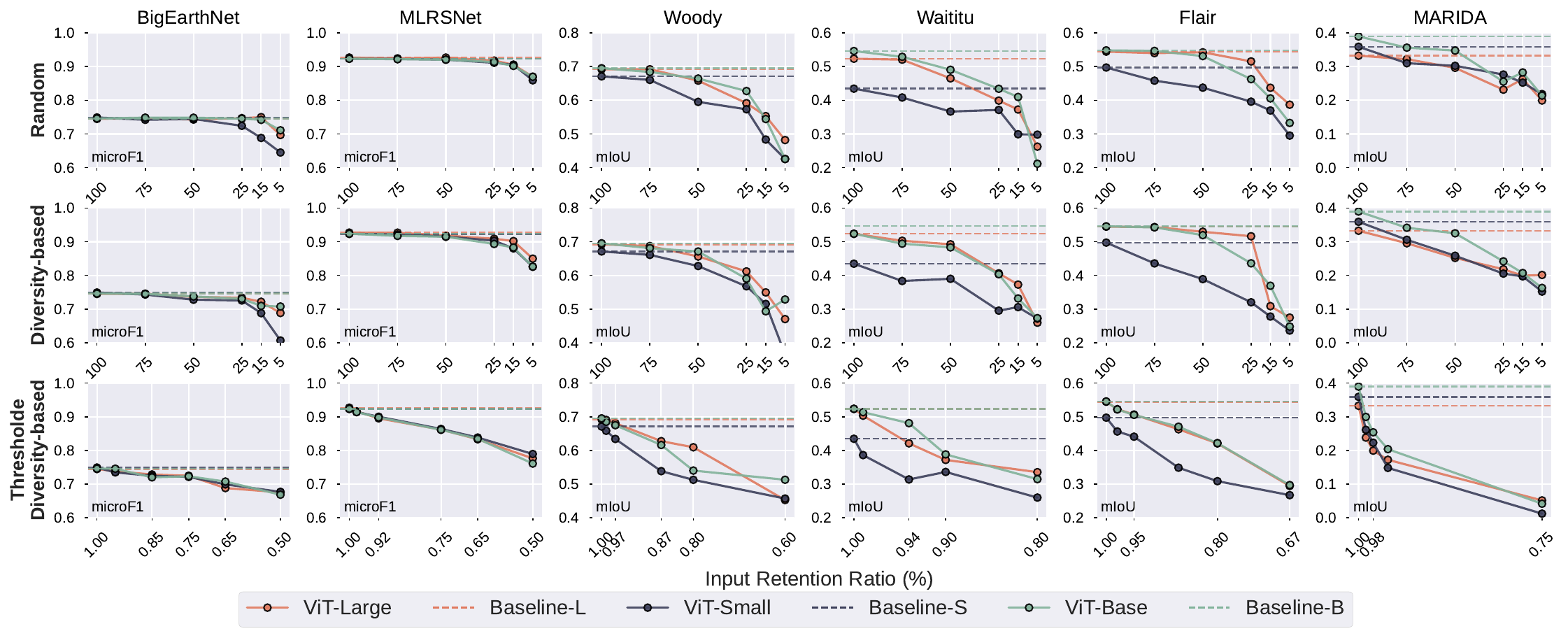}
    \caption{Examination of redundancy reduction during training across datasets. The x-axis indicates the retention ratios. For Thresholded-diversity based masking strategy we show the corresponding similarity thresholds resulting, on average, in similar retention ratios.}
    \label{fig:main_benchmark}
\end{figure*}

We study the impact of redundancy reduction using the masking strategies introduced in \cref{sec:masking_strategies}. Our investigation is structured into two experimental settings. {\textbf{\textit{Setting A}:}} Masking is applied during training with varying patch retention ratios, while models are evaluated on the full, unmasked input. This setting assesses whether models can learn sufficient task-relevant features when trained with large portions of the input removed. \textbf{\textit{Setting B}:} Masking is applied directly at inference with varying retention ratios, evaluating models trained both with and without masking. This experiment tests whether the downstream task itself can be solved using only a subset of the available 
information. 
\Cref{fig:main_benchmark} presents results for \textit{Setting A}, while \cref{fig:inference_benchmark} shows results for \textit{Setting B}. 

\textbf{\textit{Setting A: Learning under redundancy reduction.}} \Cref{fig:main_benchmark} shows that all models are able to overcome the potential object-occlusion problem at low retention ratios, with the most extreme case being BigEarthNet, where we retain almost full performance, while utilizing only $15\%$ of the input with  $\approx99.44\%$, $\approx95\%$ and $\approx94.83\%$ of the baseline for MS1, MS2 and MS3 respectively. In other words we can achieve comparable performance using $\approx9$ patches. We hypothesize that this is due to a combination of relatively low GSD along with small image size resulting in a region less than $1.2km^2$ per sample, which combined with a coarse class nomenclature focused on land use might offer strong category cues. Similarly, in MLRSNet we achieve $\approx98.93\%$ of the baseline utilizing only $25\%$ of the input for MS1 ($\approx96.71\%$ and $\approx93.24\%$ for MS2 and MS3). This is notable given that both tasks are multi-label classification with target objects distributed across the image. In semantic segmentation where exact delineation is needed, competitive performance is achieved at $50\%$ retention ratio with $\approx96.90\%$, $\approx95.22\%$, and $\approx92.65\%$ of the baseline performance for MS1, MS2 and MS3 respectively. For $75\%$ retention, we observe similar patterns in Marida, consistent performance at $98.4\%$ across MS for Woody, and consistent performance for Waititu with MS1 exhibiting a slight drop at $96.8\%$ of the baseline. 
In other words, we are able to learn diverse tasks of varying target granularities, GSDs, environmental conditions and geolocations by only utilizing the minority of the input pixels. Across all datasets, MS1 consistently matches or outperforms the informed strategies MS2 and MS3 at equivalent retention ratios.

\textbf{\textit{Setting B: Inference under redundancy reduction.}} In \cref{fig:inference_benchmark} we examine how well tasks can be solved with limited information, directly linking redundancy to the nature of the downstream application rather than the training procedure. This experiment focuses on MS1, given its simplicity and robust performance in \textit{Setting A}. Two findings emerge. First, models trained with masking are more robust at inference under low retention ratios --- an expected consequence of training under similar conditions. More importantly, models trained \textit{without} masking also maintain competitive performance under moderate masking at inference. Second, the redundancy trends observed during training are consistent at inference: multi-label classification tasks tolerate low retention ratios, while segmentation tasks are more sensitive, with performance degrading more steeply as retention ratios decrease. This tolerance is more pronounced, when masking has been applied at the training stage too. BigEarthNet remains extremely tolerant to low retention ratios even in inference, exhibiting almost full baseline performance even at $25\%$ retention ratio, in particular for models where masking was applied at the pretraining stage too (especially for $25\%$ and $15\%$ retention ratio at training time). Similarly, in MLRSNet tolerance is reduced, maintaining performance until $50\%$ retention ratio. The sensitivity of semantic segmentation tasks varies with dataset characteristics --- in MARIDA, which consists predominantly of small target classes against large sea regions, occluding even a small proportion of patches can render entire target objects invisible or their spatial continuity unrecoverable. In this case, $75\%$ retention ratio maintains performance when training with the full image. In Flair, performance is more stable in $75\%$ and $50\%$ retention ratio on inference reducing the original $1024$ patches to 512---an impressive reduction considering the quadratic cost of attention. Similarly, Waititu and Woody maintained their tolerance to $75\%$ in inference. Both datasets revolve around fine-grained tree species categories, however, their high spatial resolution and up-close point of view of UAV imagery make the tree canopies in forested regions occupy large sample portions. Notably, models trained with low retention ratios (5-15\%) show improved inference robustness at low retention ratios compared to non-masked models, but sometimes underperform at high retention ratios. 

These experiments support two conclusions. \textbf{Finding 1}: \textbf{Redundancy is an inherent property of EO imagery} --- models can learn task-relevant representations from a strict minority of input patches \textbf{across all redundancy reduction strategies (MS1, MS2, MS3)}. \textbf{Finding 2}: \textbf{EO tasks are solvable with a fraction of the available information} --- even tasks requiring spatial continuity of target classes tolerate substantial retention ratio reduction, with the degree of tolerance governed by the nature of the task and the spatial distribution and level of detail of target classes.

\textbf{On redundancy reduction strategies:} Rather than treating the comparison of masking strategies as a search for a single optimal heuristic, we frame it as a lens for understanding the sources of redundancy in EO imagery and its implications.
We observe consistent behavior across strategies, \ie, substantial portions of patches can be omitted with limited degradation in performance. The consistency of this trend 
suggests that redundancy reduction is not tightly coupled to a particular strategy, but is instead diffused across multiple aspects inherent to EO imagery. Each strategy captures a different but reasonable proxy, from assumption-free random selection to patch-level similarity-based diversity; yet all reveal the same underlying phenomenon. This robustness raises a deeper and open question: \textit{How can we identify all dimensions of redundancy?} We do not claim to answer this definitively. We demonstrate that multiple reasonable reduction methods are sufficient to exploit it, and that characterizing redundancy more formally, whether through structural properties of EO imagery such as spatiotemporal autocorrelation, or through information-theoritic measures is a promising research direction. 
In the following sections, we attempt to solidify these results by studying their generalization across targeted variations of SFs, architectural designs and distribution shifts.




\subsection{Generalization of redundancy-aware models}
\label{sec:generalization}
\begin{figure*}
    \centering
    \includegraphics[width=0.9\linewidth]{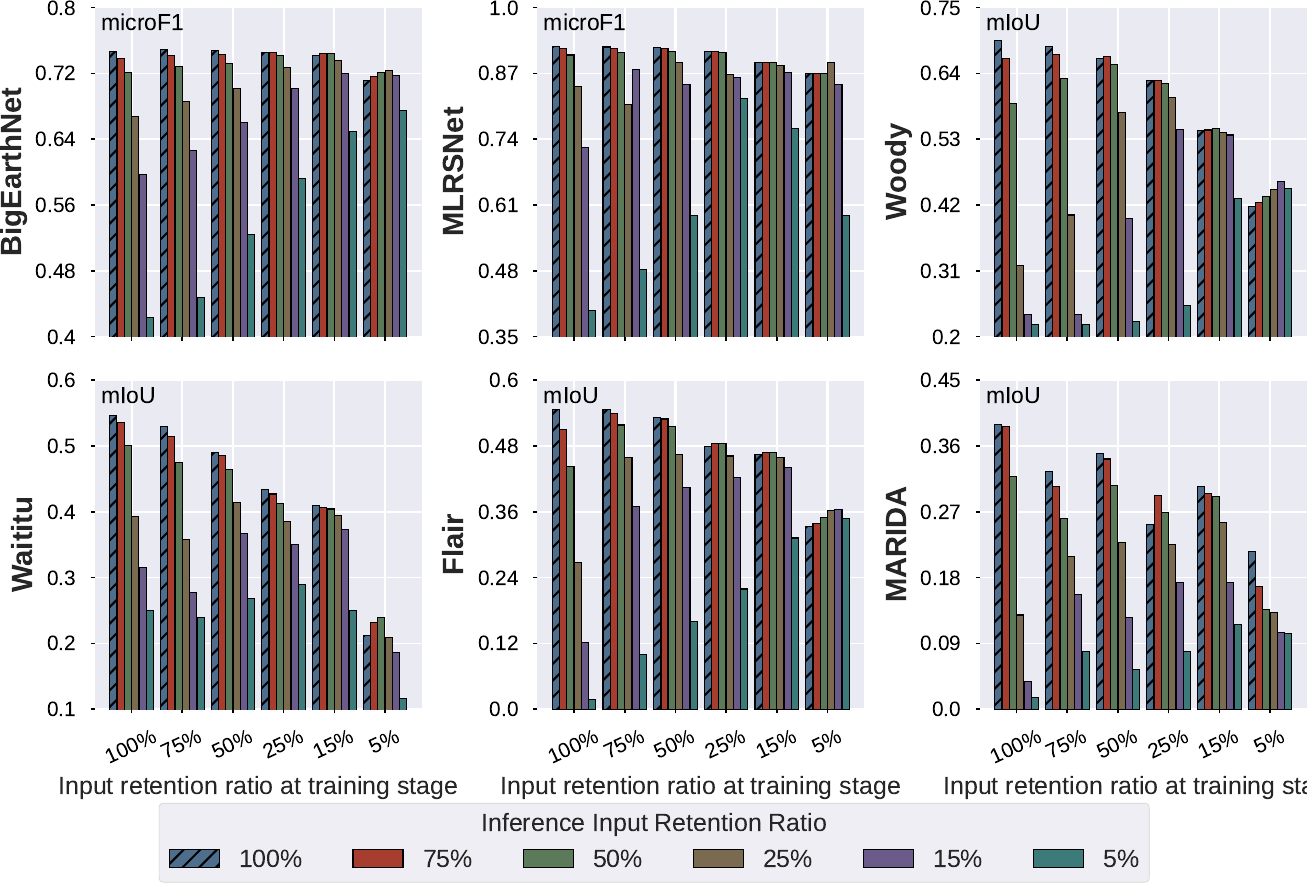}
    \caption{Examination of redundancy reduction directly on inference, for models trained with masking and without. We group models trained with the same retention ratio on the x-axis. We distinguish the inference retention ratio with color. }
    \label{fig:inference_benchmark}
\end{figure*}

\begin{figure}[h]
    \centering
    \begin{minipage}{0.49\linewidth}
        \centering
        \includegraphics[width=\linewidth]{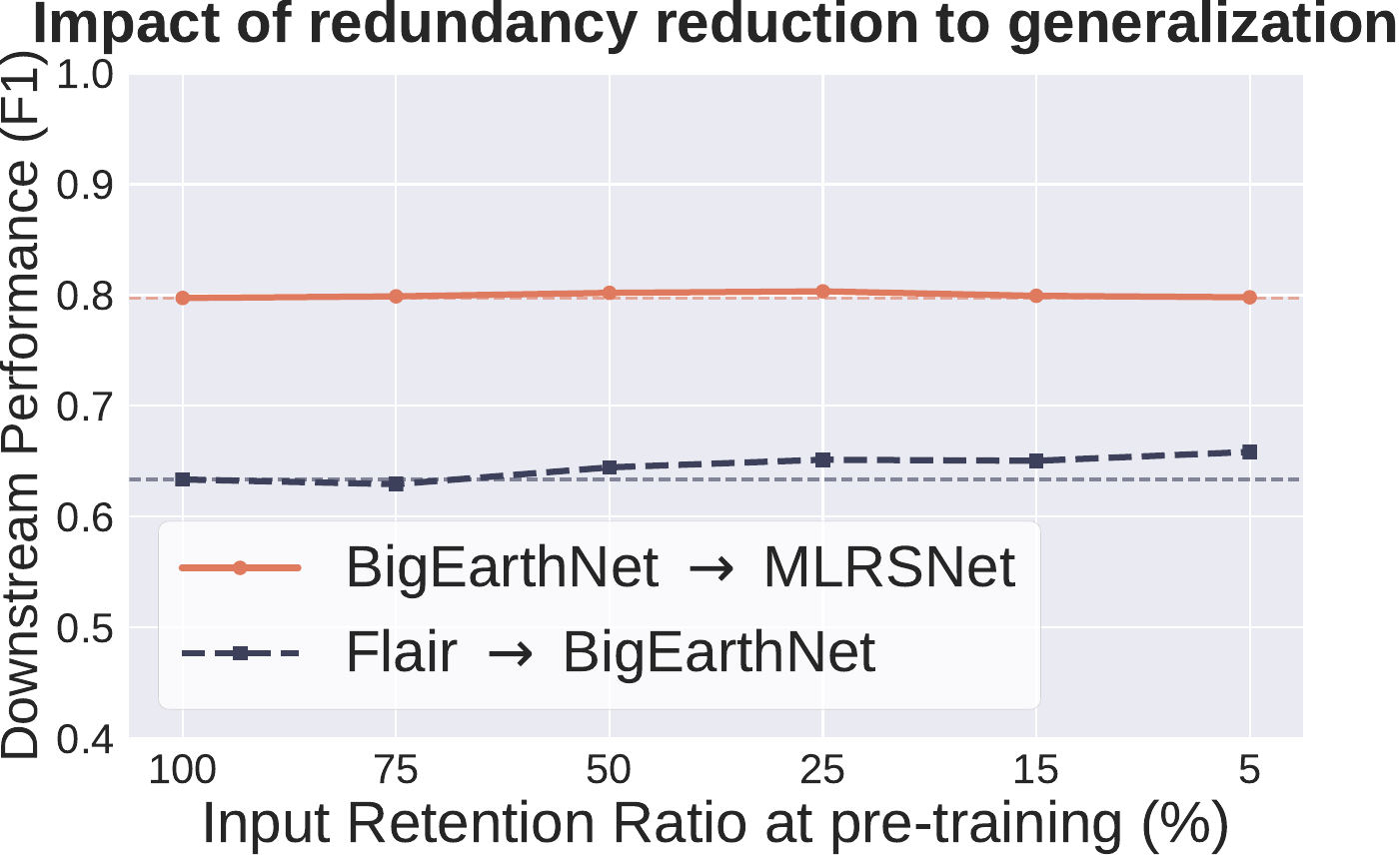}
        \caption{Linear probing of redundancy-aware models trained with varying retention ratios, to assess their generalization capacity.}
        \label{fig:generalization}
    \end{minipage}
    \hfill
    \begin{minipage}{0.49\linewidth}
        \centering
        \includegraphics[width=\linewidth]{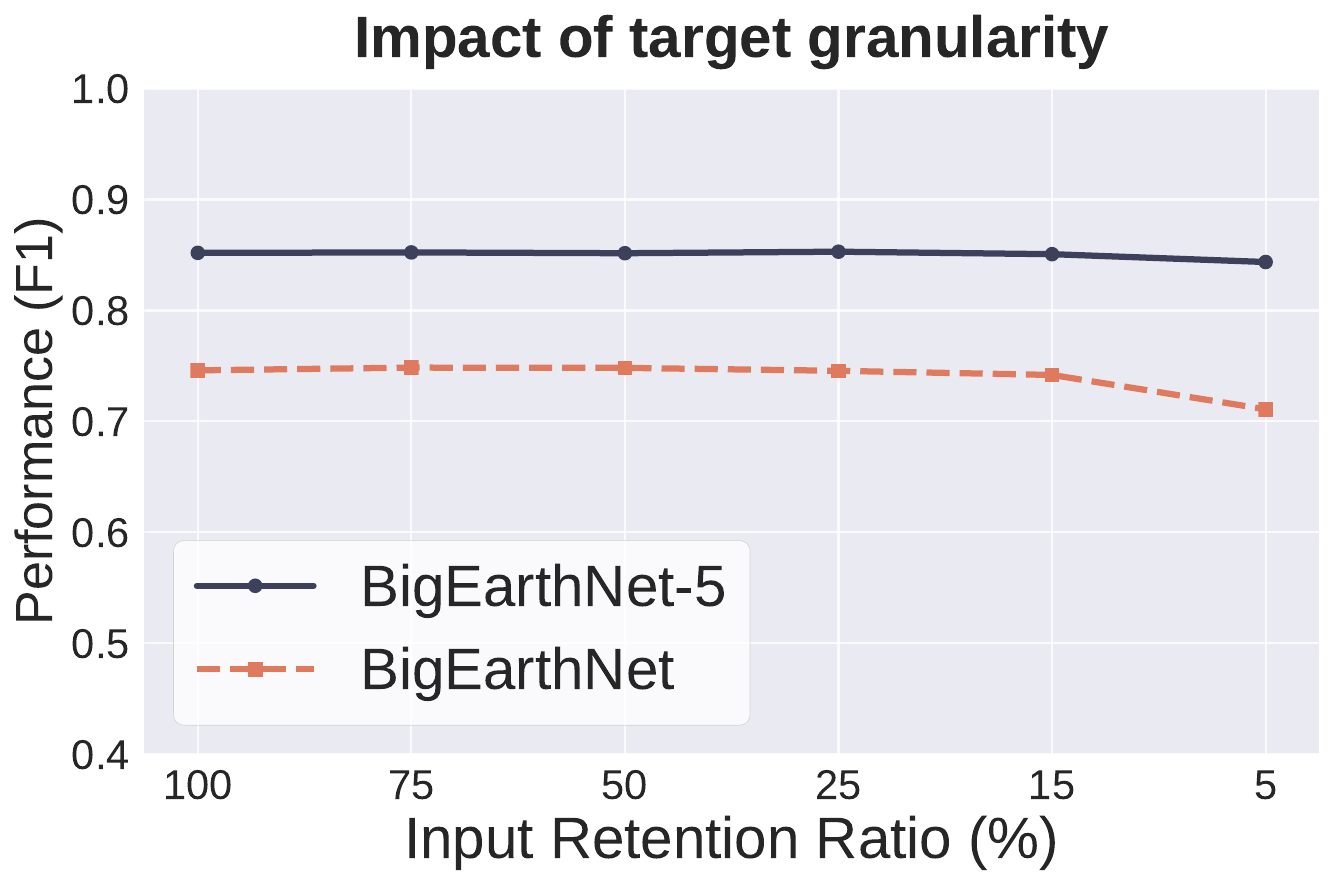}
        \caption{Examination of target granularity impact on redundancy reduction. Blue corresponds to BigEarthNet and orange to BigEarthNet-5.}
        \label{fig:target_granularity_training}
    \end{minipage}
\end{figure}

Models trained with redundancy reduction perform close to unmasked baselines using only a fraction of the input. This raises an important question: \textit{does training with reduced information affects the generalization capacity of models' representations?}  To this end, we examine how features learnt during training generalize to other tasks via linear probing. We use MS1, which performs on par with or better than diversity-based strategies (see \cref{fig:main_benchmark,fig:inference_benchmark}) while being the simplest to apply. We consider two transfer pairs designed to probe generalization across distinct SF axes.  The first pair: models pretrained on Flair and evaluated on BigEarthNet, spans SF1 (20cm to 10m GSD), and SF2 (aerial scenes in France to satellite imagery over Europe). The second pair: models pretrained on BigEarthNet and evaluated on MLRSNet, spans SF1 (10m to 10cm--10m GSD), SF2 (Sentinel-2 imagery over Europe to global multi-sensor scenes), and SF3 (coarse land cover categories to a dataset ranging from fine-grained objects such as \textit{crosswalk} to broad scene-level classes such as \textit{sea}). 
We use ViT-Base focusing on the RGB channels for this experiment, which are common across datasets. \Cref{fig:generalization} summarizes this experiment. Remarkably, transfer performance is consistently similar to or slightly better than the full-input baseline across masking ratios, even at extreme retention ratios. \textbf{Finding 3: Redundancy-aware training does not compromise the quality of learned representations} --- redundancy-aware training retains the quality of learned representations, and may in fact improve their transferability by encouraging the model to capture the most discriminative and general features in the scene.

\subsection{Impact of Ground Sampling Distance}
\begin{figure}
    \centering
    \includegraphics[width=\linewidth]{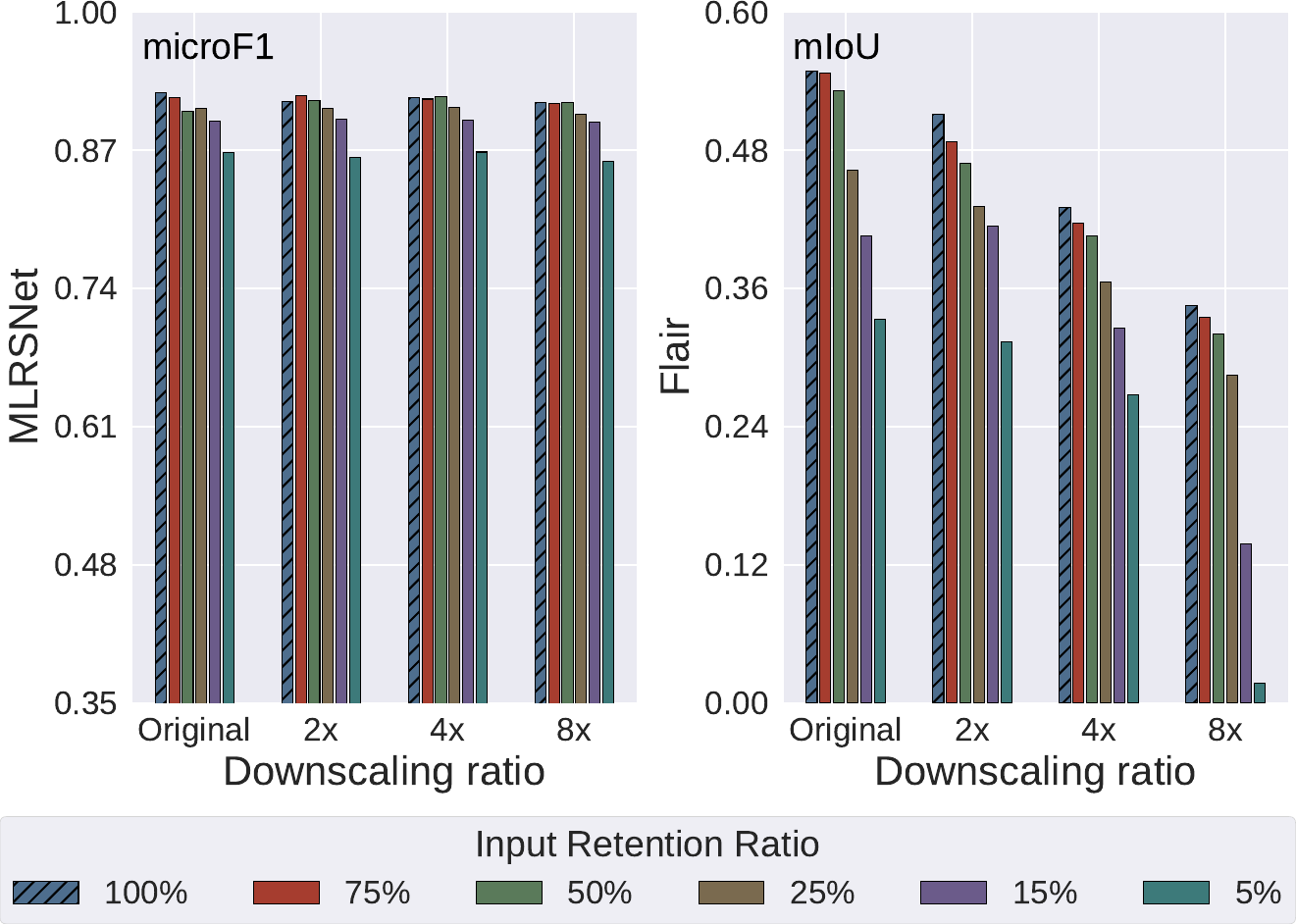}
    \caption{Examination of GSD impact on the robustness of redundancy reduction for MLRSNet and Flair.}
    \label{fig:downscaling_effect}
\end{figure}
\Cref{fig:main_benchmark,fig:inference_benchmark} demonstrated that redundancy reduction is robust across datasets spanning a diverse range of GSDs. Yet, each dataset involves a task with unique characteristics that may confound this robustness. Here we attempt to isolate GSD as a factor and study how redundancy exploitability varies with spatial resolution. To this end, we select two datasets, MLRSNet and Flair, and perform progressive downscaling, assessing how performance under \textit{Setting A} varies across resolutions. MLRSNet proves remarkably robust to downsampling --- even an extreme resolution reduction from $224\times224$ (10cm--10m GSD) to $32\times32$ (80cm--80m GSD) results in only marginal performance degradation. Crucially, the relative impact of redundancy reduction remains consistent across all tested resolutions. The same pattern holds for Flair, downscaled from $512\times512$ (20cm GSD) to $64\times64$ (1.6m GSD): while the task becomes progressively more challenging at lower resolutions, the redundancy reduction trends are preserved. \textbf{Finding 4: Redundancy in EO imagery is consistent across GSD variations} --- the relative impact of redundancy reduction on performance is stable regardless of spatial resolution, suggesting that redundancy is not an artifact of a specific spatial resolution but a consistent property across the SF1 axis of EO variability.
\subsection{Impact of target granularity}
\begin{figure}
    \centering
    \includegraphics[width=\linewidth]{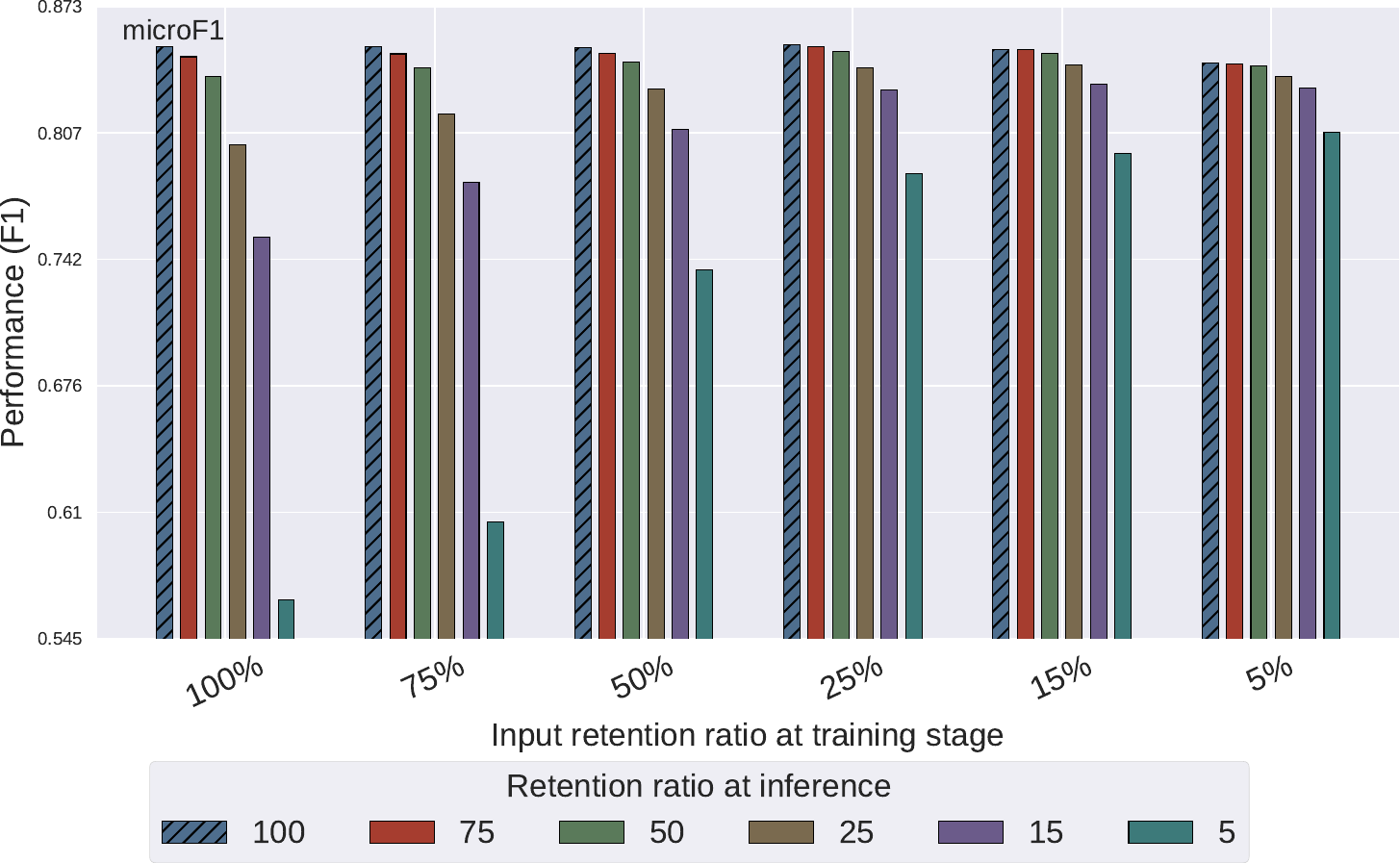}
    \caption{Impact of redundancy reduction on inference for BigEarthNet-5.}
    \label{fig:target_granularity_inference}
\end{figure}

Our benchmark (see \cref{fig:main_benchmark,fig:inference_benchmark}) shows that redundancy reduction has a consistent effect across tasks. Since the masking strategies presented in \cref{sec:masking_strategies} are task-agnostic and different datasets exhibit different target granularities, we hypothesize that this consistency extends across granularities as well. To validate this, we isolate the SF3 (target granularity) utilizing BigEarthNet-5 (see \cref{sec:datasets}), and examine the behavior of redundancy reduction with MS1 across retention ratios comparing against the corresponding models trained on the full BigEarthNet dataset. We do that for both settings presented in \cref{sec:existence}, \ie, masking only during training (\textit{Setting A}), and masking during inference across training setups (\textit{Setting B}). The results of these experiments are summarized in \cref{fig:target_granularity_training,fig:target_granularity_inference} for training and inference examination respectively. In this case the exact performance scores between the two datasets are not relatable, as BigEarthNet-5 is clearly an easier task. Nevertheless, \cref{fig:target_granularity_training} shows that both datasets exhibit the same performance reduction trend as information occlusion becomes more aggressive \ie full performance maintanance until $15\%$ input retention and then a slight decline. Similarly, in \cref{fig:target_granularity_inference} we observe a similar trend in perfromance reduction, compared to full BigEarthNet presented in \cref{fig:inference_benchmark}, across inference retention ratios for all training setups. All the above leads us to the following conclusion \textbf{Finding 5: The impact of redundancy reduction is consistent across target granularities}, making the results of this study applicable to a broad range of EO applications.


\subsection{Efficiency-Performance tradeoff}
\begin{figure}
    \centering
    \includegraphics[width=\linewidth]{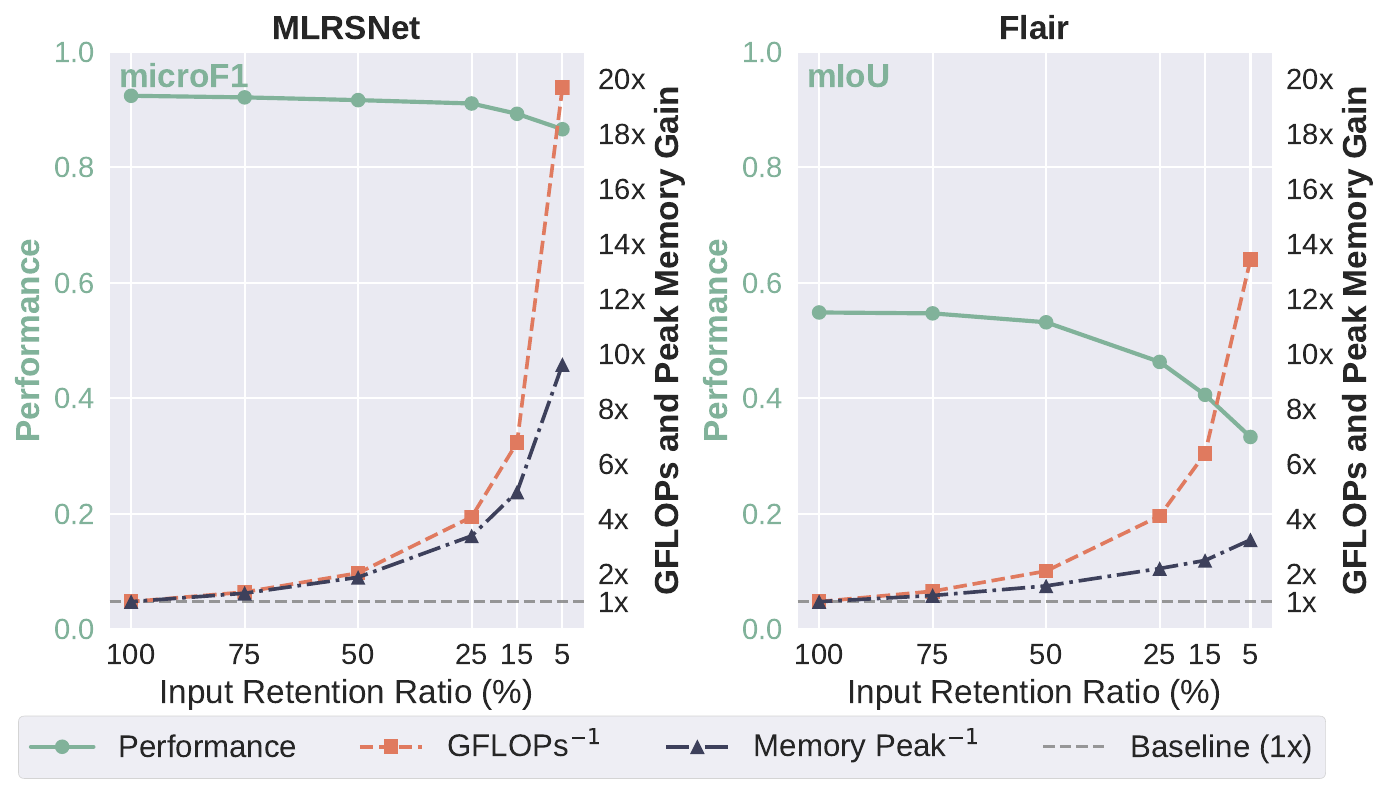}
    \caption{Performance-Efficiency trade-off in training.}
    \label{fig:train_gains}
    \includegraphics[width=\linewidth]{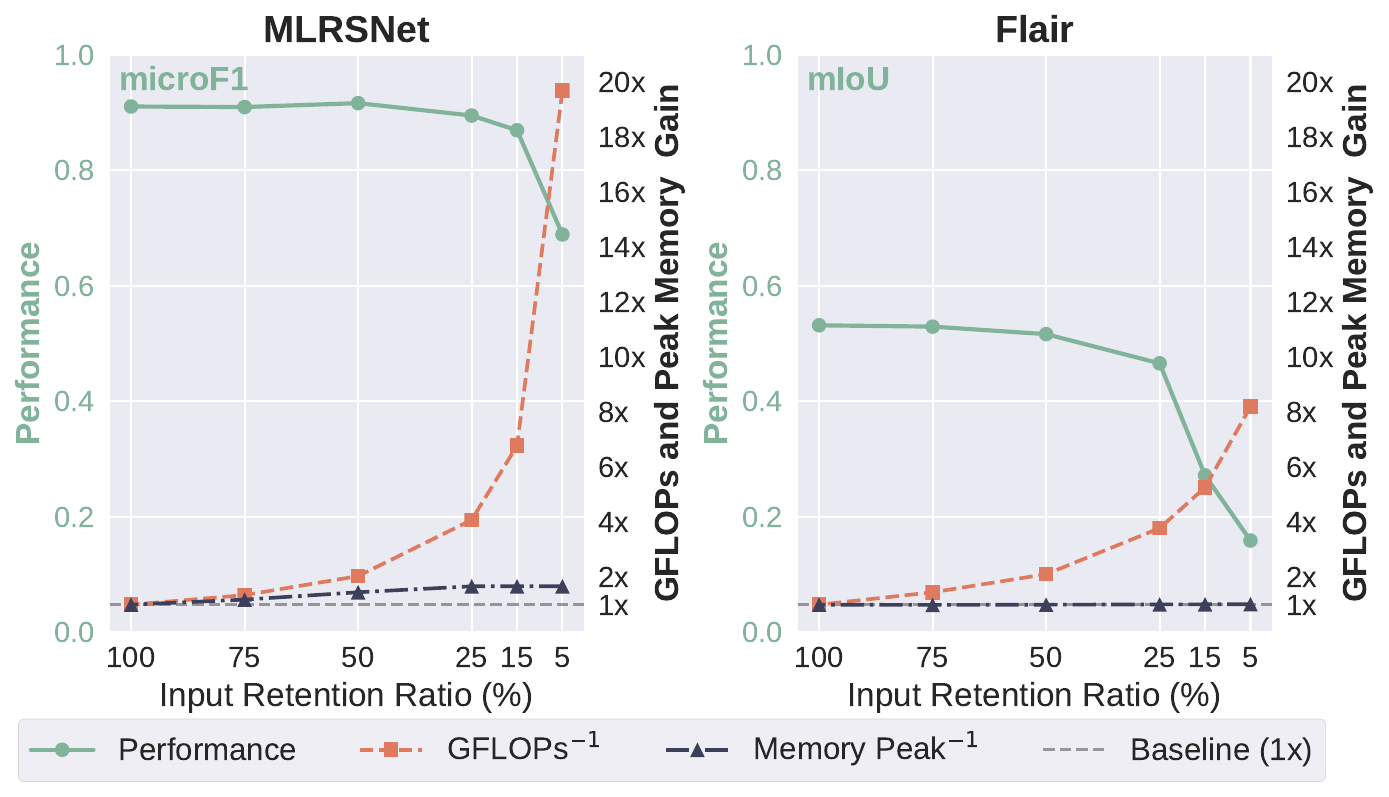}
    \caption{Performance-Efficiency trade-off in inference.}
    \label{fig:inference_gains}
\end{figure}

Redundancy reduction directly translates to computational savings by processing fewer input patches. We examine these gains across retention ratios for training and inference using a ViT-Base with MS1 on MLRSNet and Flair --- the highest resolution datasets in our benchmark at $224\times224$
and $512\times512$ pixels respectively. Results are summarized in \cref{fig:train_gains,fig:inference_gains}. For MLRSNet, retaining $25\%$
 of the input preserves near-full predictive performance while yielding a $\approx4\times$ reduction in GFLOPs and memory utilization. At $5\%$ retention, these gains scale to $20\times$ and $10\times$ respectively, at a performance cost of $\approx6\%$. For Flair, the efficiency-performance sweet spot shifts to $50\%$ retention, yielding $\approx2\times$ savings in GFLOPs and memory. These gains could be further amplified by designing a decoder that fully omits masked patches, a current limitation of RViT-UPerNet. The same patterns hold at inference (see \cref{fig:inference_gains}), underscoring the potential of redundancy reduction for edge deployment.
\subsection{Consistency across design choices}
\begin{figure}
    \centering
    \includegraphics[width=0.9\linewidth]{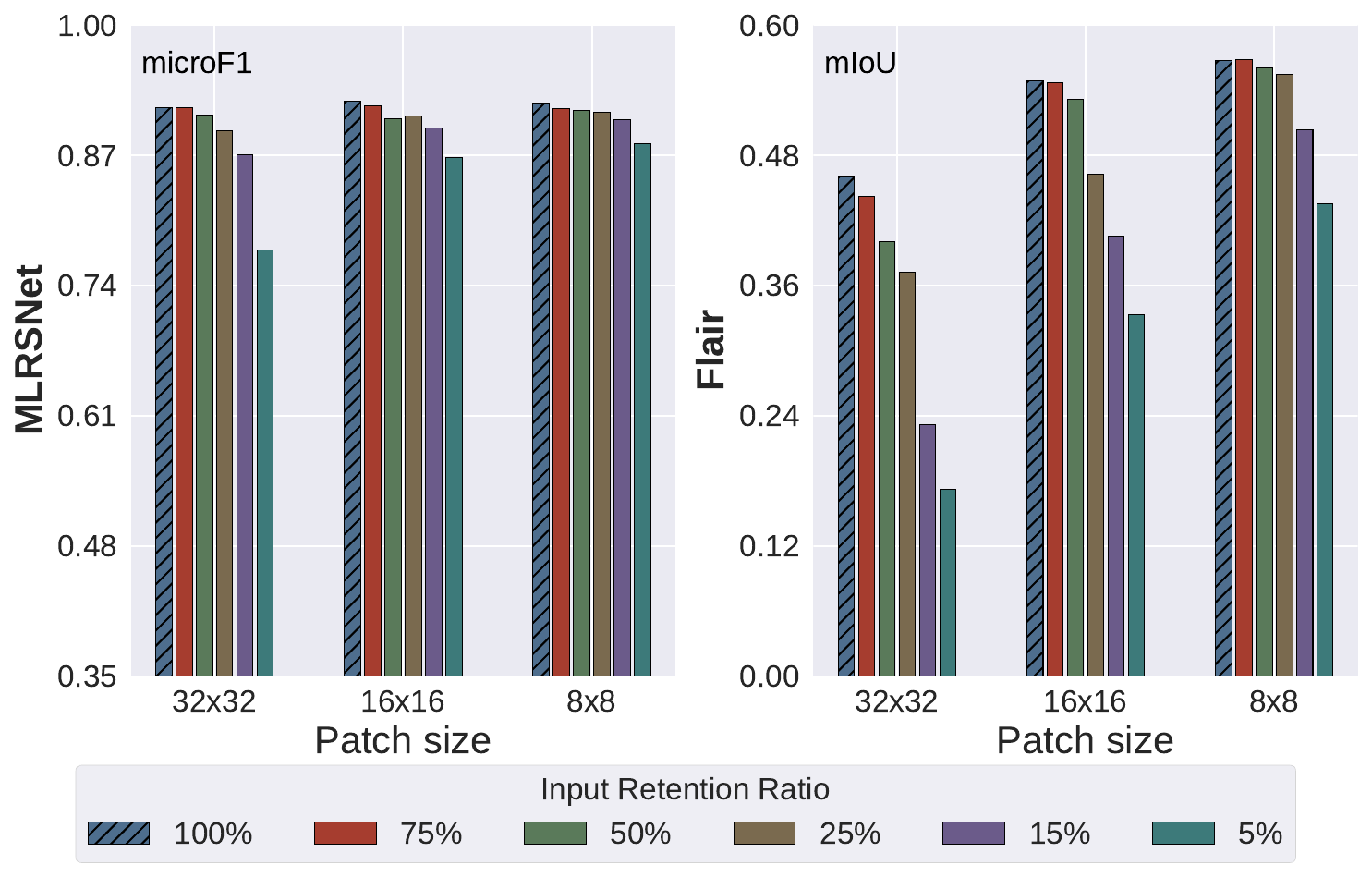}
    \caption{Impact of Transformer's patch size on redundancy phenomenon.}
    \label{fig:patch_size}
\end{figure}

Our investigation has so far used a fixed patch size of $16\times16$. As redundancy is reducted based on patches, this could have implications on the results of our investigation. We evaluate the impact of varying patch size using a ViT-Base with MS1 on MLRSNet and Flair. As shown in \cref{fig:patch_size}, while performance consistently improves with smaller patch sizes (resulting in a higher number of patches), the redundancy reduction trends remain stable across all configurations. This suggests that redundancy exploitability is independent of patch size, and that finer-grained patch sizes can be exploited at only a fraction of the computational and memory cost.  
\section{Conclusion \& Future Work}
In this work, we presented a systematic framework for studying redundancy in EO imagery, demonstrating that it is both substantial and pervasive across tasks, sensors, GSDs, geolocations, and architectural designs. We examined three redundancy reduction strategies that achieve comparable performance to unreduced baselines ($\approx98.5\%$ of baseline) at a fraction of the computational cost ($\approx4\times$ fewer GFLOPs) during both training and inference. These efficiency gains carry direct implications for scalable pretraining, edge inference, and foundation model development. Our findings open several promising research directions, including the study of temporal and cross-channel redundancy, the design of decoders that fully exploit patch-level masking, and the formal characterization and isolation of redundancy factors. We hope this work motivates redundancy-aware architectures and training strategies tailored to the structural properties of EO imagery.
{
    \small
    \bibliographystyle{ieeenat_fullname}
    \bibliography{main}
}

 \clearpage
\twocolumn[{
  \begin{center}
    {\LARGE\textbf{Supplementary Material}}
  \end{center}
}]
\setcounter{section}{0}
\renewcommand{\thesection}{A\arabic{section}}
\renewcommand{\sectionautorefname}{Suppl.}
\section{Visualization of masking strategies}
\label{sec:viz_mask}
\begin{figure*}
    \centering
    \includegraphics[width=1\linewidth]{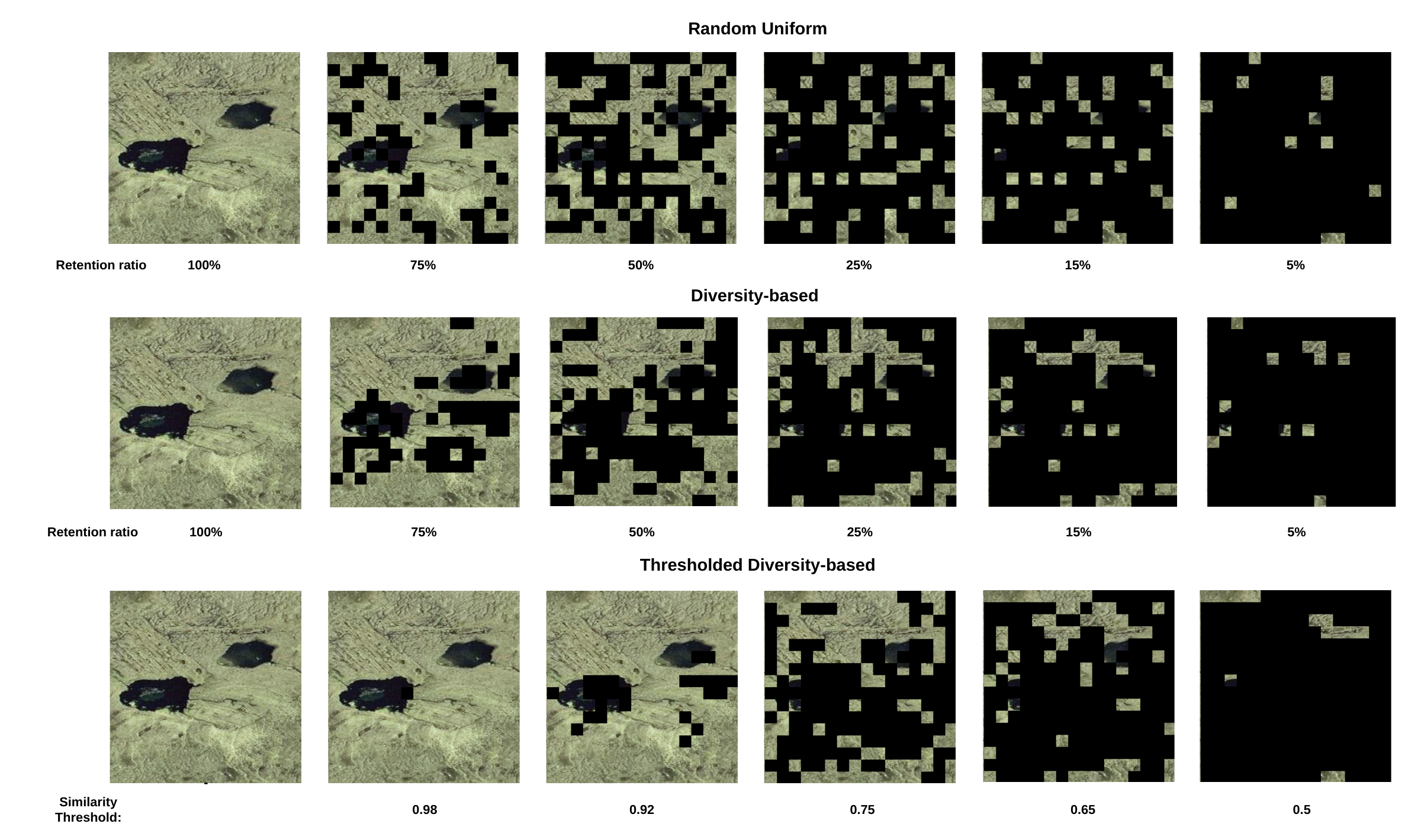}
    \caption{Visual examples of the masking strategies at varying retention ratios on the MLRSNet dataset}
    \label{fig:masking_strategies_Visualization}
\end{figure*}

\Cref{fig:masking_strategies_Visualization} presents visual examples of the masking strategies at varying retention ratios on the MLRSNet dataset. Notably, the Thresholded Diversity-based strategy first occludes the most semantically similar regions, in this case a lake, before progressing to the rest of the image. As retention ratios decrease, all strategies converge in behavior, with Random Uniform maintaining a more spatially balanced distribution, as expected.

\section{Threshold-Retention ratio correspondence}
\label{sec:suppl_correspondence}
\Cref{tab:correspondence} showcases the correspondence of similarity thresholds to retention ratios per dataset for the Thresholded Diversity-based masking strategy.
\begin{table}[h]
\label{tab:correspondence}
\centering
\caption{Correspondence of similarity thresholds to retention ratios per dataset}
\begin{tabular}{lccccc}
\toprule
 & \multicolumn{5}{c}{\textbf{Retention ratio\%}} \\
\cmidrule(lr){2-6}
\textbf{Dataset} & \textbf{75} & \textbf{50} & \textbf{25} & \textbf{15} & \textbf{5} \\
\midrule
BigEarthNet & 0.95 & 0.85 & 0.75 & 0.65 & 0.50 \\
MLRSNet     & 0.98 & 0.92 & 0.75 & 0.65 & 0.50 \\
FLAIR       & 0.98 & 0.95 & 0.87 & 0.80 & 0.67 \\
Woody       & 0.99 & 0.97 & 0.87 & 0.80 & 0.60 \\
Waititu     & -    & 0.99 & 0.94 & 0.90 & 0.80 \\
MARIDA      & -    & 0.99 & 0.98 & 0.96 & 0.75 \\
\bottomrule
\end{tabular}
\end{table}

\end{document}